\documentclass[10pt,twocolumn,letterpaper]{article}

\usepackage[pagenumbers]{cvpr} % To force page numbers, e.g. for an arXiv version

\definecolor{cvprblue}{rgb}{0.21,0.49,0.74}
\usepackage[pagebackref,breaklinks,colorlinks,allcolors=cvprblue]{hyperref}

\def\confName{CVPR}
\def\confYear{2027}

\title{CH-Cache: Horizon-Aware Chebyshev-Hermite Feature Fusion for Training-Free Diffusion Acceleration}

\author{
\textbf{Hongzhe Wu}$^{1,*}$,
\textbf{Junxiang Qiu}$^{2,*}$,
\textbf{Shuo Wang}$^{2}$,
\textbf{Lin Liu}$^{2}$,
\textbf{Zhenhua Tang}$^{3}$,
\textbf{Hui Ma}$^{1}$,
\textbf{Yanbin Hao}$^{1,\dagger}$\\[3pt]
{\small
$^{1}$Hefei University of Technology,
$^{2}$University of Science and Technology of China,
$^{3}$University of Macau
}\\[2pt]
{\small
2025110468@mail.hfut.edu.cn,
\{qiujx, liulin0725\}@mail.ustc.edu.cn,
shuowang.edu@gmail.com
}\\
{\small
zhenhuat@foxmail.com,
\{huima,haoyanbin\}@hfut.edu.cn
}\\[2pt]
{\small $^{*}$Equal contribution.
\quad $^{\dagger}$Corresponding author.}
}

\usepackage{cuted}
\usepackage{caption}
\usepackage{bm}
\usepackage[table]{xcolor}
\usepackage{algorithm}
\usepackage{algpseudocode}
\usepackage{amsmath}
\usepackage{amssymb}
\definecolor{oursgray}{RGB}{242,242,242}

\begin{document}
\maketitle
\begin{strip}
    \centering
    \includegraphics[width=\textwidth]{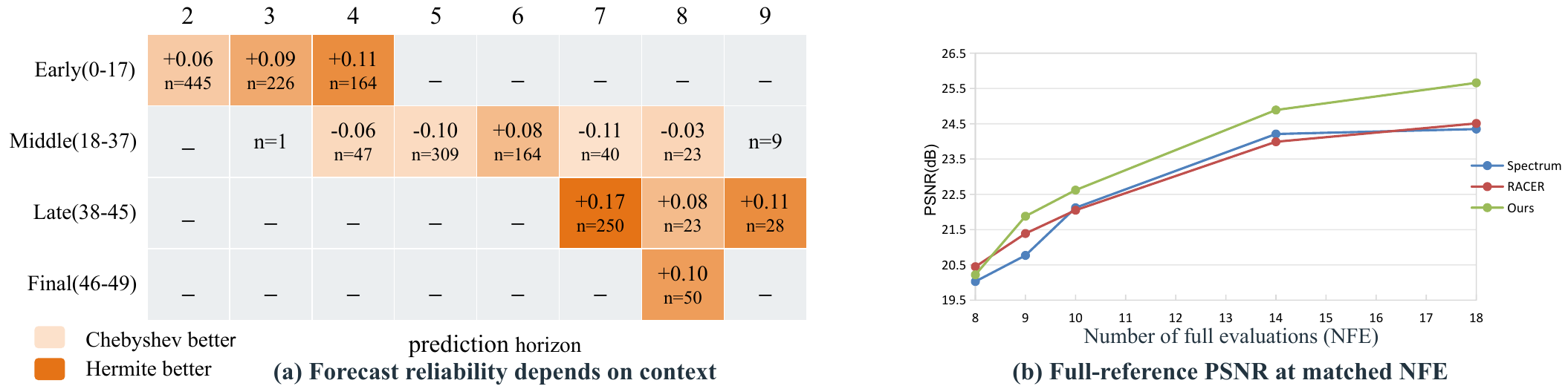}
    \captionsetup{hypcap=false}
    \captionof{figure}{\textbf{Forecasting reliability and reconstruction fidelity on FLUX.1-dev.}
    (a) Normalized error advantage $d=(E_C-E_H)/(E_C+E_H+\epsilon)$ between Chebyshev (C) and Hermite (H) forecasts across prediction horizons and denoising phases at NFE$=14$. Positive values favor H; negative values favor C. Gray cells contain fewer than 20 Full-step observations.
    (b) Full-reference PSNR on 200 DrawBench prompts. Our budget-specific Chebyshev-Hermite (C/H) fusion achieves the highest PSNR among the plotted methods at four of five budgets.}
    \captionsetup{hypcap=true}
    \label{fig:motivation}
\end{strip}

\begin{abstract}
Diffusion Transformers (DiTs) achieve high-quality image generation but incur substantial computational costs due to repeated network evaluations. Feature caching reduces this cost by reusing or forecasting intermediate features across denoising steps. However, existing forecasting methods often rely on individual predictors or fixed approximation strategies, overlooking the varying reliability of different forecasters across prediction horizons and denoising stages. Even methods employing multiple forecasters, such as RACER, primarily exploit their disagreement to regulate trust in a base prediction and trigger cache refreshes, rather than combining their complementary predictions. To address these limitations, we propose \textbf{Chebyshev-Hermite Cache (CH-Cache)}, a training-free framework with three key components: \textbf{(1) Dual-basis feature forecasting}, which combines Chebyshev and Hermite predictors to exploit their complementary approximation capabilities; \textbf{(2) Horizon-aware adaptive fusion}, which leverages relative prediction errors observed at existing Full steps to update fusion weights across nearby prediction horizons, enabling online reliability adaptation without additional network evaluations; and \textbf{(3) Disagreement-guided budgeted refresh}, which uses intermediate-layer predictor disagreement to guide adaptive cache refreshing, together with a refresh-and-repay mechanism that preserves the prescribed computation budget. Experiments on FLUX.1-dev and Stable Diffusion 3.5-Large across five computation budgets demonstrate the effectiveness of our framework. On FLUX.1-dev, our method improves PSNR by 0.57 dB (2.6\%) over RACER at NFE$=10$. On SD3.5-Large, it achieves average PSNR gains of 6.29 dB (63.8\%) and 5.85 dB (56.8\%) over TaylorSeer and HiCache, respectively.
\end{abstract}    
\section{Introduction}
\label{sec:intro}

Diffusion Transformers (DiTs) have demonstrated remarkable capabilities in high-fidelity image generation, powering text-to-image models such as FLUX.1 and Stable Diffusion 3~\cite{peebles2023scalable,flux2024,esser2024scaling}. However, their inference remains computationally expensive. Generating a single image typically involves dozens of sequential denoising steps, each requiring the execution of costly Transformer blocks. The resulting latency can hinder the practical deployment of large-scale diffusion models. Consequently, accelerating diffusion inference without compromising generation quality has become an important research direction.

Existing diffusion acceleration methods mainly follow two directions. The first reduces the number of denoising steps through efficient numerical solvers, improved sampling strategies, or model distillation~\cite{song2020denoising,lu2022dpm,luo2023latent}. The second reduces the computational cost of individual denoising steps by exploiting temporal redundancy in intermediate features~\cite{ma2024deepcache,selvaraju2024fora}. In particular, training-free feature caching methods perform full Transformer computations at selected timesteps (Full steps) and approximate intermediate features at the remaining steps (Cache steps). Early approaches reuse previously computed features~\cite{ma2024deepcache,selvaraju2024fora}, while more recent methods forecast how these features evolve rather than directly reusing past values. For example, TaylorSeer predicts the evolution of features using Taylor expansions~\cite{liu2025reusing}, while Spectrum employs a Chebyshev polynomial approximation to capture feature trajectories~\cite{han2026adaptive}, and HiCache introduces scaled Hermite forecasting~\cite{feng2026hicache}. As full computations are replaced by feature approximations at selected steps, the reliability of these forecasts becomes important for preserving generation quality.

A central challenge is that feature forecasters may differ in reliability across prediction horizons and denoising stages. Extrapolation based on recent feature variations may capture short-term changes but become less reliable at longer horizons. Polynomial approximation using historical features, such as the Chebyshev approach in Spectrum~\cite{han2026adaptive}, may capture broader temporal trends but can respond differently to localized variations. These differences suggest that a single forecasting strategy may not remain preferable throughout the denoising trajectory. Although RACER employs multiple forecasters, it primarily uses their disagreement to adjust a base prediction and guide cache refresh decisions, rather than explicitly adapting the contributions of their forecast outputs based on observed prediction errors~\cite{li2026disagree}. This motivates investigating whether relative predictor reliability can be used more directly to construct feature forecasts.

To investigate this issue, we analyze the prediction errors of Chebyshev and Hermite feature forecasters at existing Full evaluation steps. As shown in Fig.~\ref{fig:motivation}(a), on FLUX.1-dev at NFE$=14$, the observed relative accuracy of the two predictors varies across prediction horizons and denoising phases. These observations suggest that consistently favoring either forecaster may not be optimal throughout the denoising process. Meanwhile, at eligible Full steps, features computed at the current accelerated latent can be compared with predictions constructed from preceding Full-step features, enabling direct evaluation of forecasting errors without additional denoiser evaluations. These findings motivate a key question: \emph{Can such observed errors be used to adaptively determine how much each forecaster should contribute to future predictions?}

Motivated by these observations, we propose \textbf{Chebyshev-Hermite Cache (CH-Cache)}, a training-free framework that adaptively fuses Chebyshev and Hermite feature forecasts during diffusion sampling. At eligible Full steps, CH-Cache compares pre-update forecasts from both predictors with computed features to estimate their relative reliability. The resulting reliability target is used to update horizon-indexed fusion weights, with stronger feedback transferred to nearby horizons to account for the sparsity of Full-step observations. In addition, for cache refresh control, we retain RACER's budget-preserving refresh-and-repay mechanism and use Chebyshev-Hermite disagreement at a selected intermediate layer to guide refresh decisions. CH-Cache therefore combines online reliability feedback with budget-preserving refresh control, without requiring additional model training or Full denoiser evaluations beyond the prescribed budget.

We evaluate CH-Cache on FLUX.1-dev and SD3.5-Large across five matched budgets. On FLUX.1-dev, our approach improves full-reference reconstruction fidelity over representative feature caching baselines, achieving the best SSIM and LPIPS across all five evaluated budgets and the highest PSNR in four of them (Fig.~\ref{fig:motivation}(b)). On SD3.5-Large, CH-Cache also achieves the best SSIM and LPIPS among the compared methods at all five budgets, although its PSNR advantage is not consistent across budgets. The results highlight backbone-dependent differences in forecasting behavior, motivating further analysis of predictor combinations, online feedback and fusion sensitivity. Together, these experiments investigate the effectiveness and limitations of horizon-aware adaptive fusion for efficient diffusion inference. 

Our contributions are summarized as follows:
\begin{itemize}
    \item \textbf{Horizon-dependent forecasting reliability.} We investigate the complementary behavior of Chebyshev and Hermite feature forecasters and analyze how their relative prediction accuracy varies across forecasting horizons and computational budgets.
    \item \textbf{Feedback-guided adaptive predictor fusion.} We develop a training-free feature forecasting framework that leverages prediction errors observed at existing Full steps to update fusion weights online. A horizon-aware feedback mechanism enables differentiated weight adaptation without additional denoiser evaluations.
    \item \textbf{Evaluation under matched computation budgets.} We conduct experiments on FLUX.1-dev and SD3.5-Large across multiple acceleration levels, demonstrating improved reconstruction fidelity on FLUX.1-dev and analyzing the influence of predictor combinations, online feedback, and fusion hyperparameters.
\end{itemize}

\section{Related Work}
\label{sec:related_work}

\noindent\textbf{Diffusion sampling acceleration.}
Diffusion models require multiple sequential network evaluations, motivating extensive research on efficient inference. One line of work reduces the number of denoising steps through improved sampling strategies. DDIM~\cite{song2020denoising} introduces non-Markovian sampling, while DPM-Solver~\cite{lu2022dpm,lu2025dpm,zheng2023dpm} employs high-order numerical solvers. Flow matching~\cite{lipman2022flow} and rectified flow~\cite{liu2022flow,esser2024scaling} formulate generative modeling through learned velocity fields, underpinning modern flow-based diffusion Transformers. Consistency models~\cite{song2023consistency} enable few-step generation through training or distillation~\cite{qiu2025multimodal,salimans2022progressive,yin2024one}. Feature caching offers another route by reusing intermediate computation within a pretrained model~\cite{selvaraju2024fora}. Our work follows this training-free direction: it retains the sampling time grid and model parameters while replacing selected Transformer evaluations with forecasts from previously computed features.

\noindent\textbf{Feature caching.}
Feature caching exploits temporal redundancy in denoising trajectories. DeepCache~\cite{ma2024deepcache} reuses high-level U-Net features while updating lower-level features, whereas subsequent approaches further exploit redundancy across blocks, layers and tokens ~\cite{selvaraju2024fora,ma2024learning,zou2024accelerating,zheng2025compute}. Error-compensated approaches further refine cached features~\cite{qiu2025eoc,qiu2025goc}. Moving from reuse to extrapolation, TaylorSeer~\cite{liu2025reusing} predicts features using truncated Taylor expansions, while FoCa~\cite{zheng2026forecast} introduces feature-ODE forecasting and calibration. Spectrum~\cite{han2026adaptive} models feature trajectories with Chebyshev polynomials and fits their coefficients through ridge regression. HiCache~\cite{feng2026hicache} introduces scaled Hermite forecasting with a dual-scaling mechanism to improve numerical stability. Other approaches explore frequency-, subspace-, and token-adaptive feature approximation~\cite{liu2025freqca,chen2026forecast,zhu2026tap}. Building on the Chebyshev and Hermite forecasts, CH-Cache uses prediction errors observed at existing Full evaluations to update fusion weights for subsequent Cache steps.

\noindent\textbf{Adaptive cache control.}
Adaptive caching also requires deciding when a forecast should be replaced by a fresh computation. TeaCache~\cite{liu2025timestep} and AdaCache~\cite{kahatapitiya2025adaptive} adapt caching based on input or feature variations, while confidence-gated forecasting~\cite{guan2025forecasting} and SpeCa~\cite{liu2025speca} use prediction-related signals to guide cache decisions. Other approaches further explore sensitivity-aware, globally optimized, or budget-aware cache scheduling~\cite{haghighi2026sencache,zhang2026softcap,cui2026denoising}. RACER~\cite{li2026disagree} uses forecaster disagreement for prediction shrinkage and budget-preserving cache refreshing. CH-Cache retains RACER's budget-preserving refresh-and-repay mechanism while using Chebyshev–Hermite disagreement as the refresh signal.
\section{Method}
\label{sec:method}

\subsection{Preliminaries and Problem Setup}
\label{sec:preliminaries}

\noindent\textbf{Diffusion inference and feature caching.}
We consider inference with a pretrained diffusion Transformer~\cite{peebles2023scalable} over $T$ denoising steps, indexed by $t=0,\ldots,T-1$ in execution order. Let $\bm{x}_t$ denote the latent at step $t$, $\sigma_t$ the corresponding noise level, and $c$ the text condition. Under an Euler flow sampler~\cite{esser2024scaling}, the latent is updated as
\begin{equation}
 \bm{x}_{t+1}=\bm{x}_t+(\sigma_{t+1}-\sigma_t)\bm{v}_\theta(\bm{x}_t,\sigma_t,c).
 \label{eq:ch-sampling}
\end{equation}
A Full evaluation of $\bm{v}_\theta$ executes the expensive Transformer blocks. Feature caching~\cite{ma2024deepcache,selvaraju2024fora} reduces this cost by forecasting intermediate activations from previously computed features.

Let $\bm{F}_t^{(\ell)}$ denote the image-token output at block $\ell$, and let $L$ be the last cached block. A Full step computes all blocks and stores their actual features as anchors, whereas a Cache step predicts $\bm{F}_t^{(L)}$ from these anchors and forwards the prediction through the unchanged output head. Throughout the paper, an actual feature is computed at the current accelerated latent $\bm{x}_t$, not obtained from a separate full-sampling reference trajectory. Let $\mathcal{A}_t=\{a_0,\ldots,a_j\}$ denote the indices of available Full anchors before step $t$, with $a_j<t$ being the most recent anchor. We define the prediction horizon as $h_t=t-a_j$. All forecasts are constructed exclusively from previously computed Full-step features.

\begin{figure*}[t]
 \centering
 \IfFileExists{figures/method_overview.pdf}{%
   \includegraphics[width=0.98\textwidth]{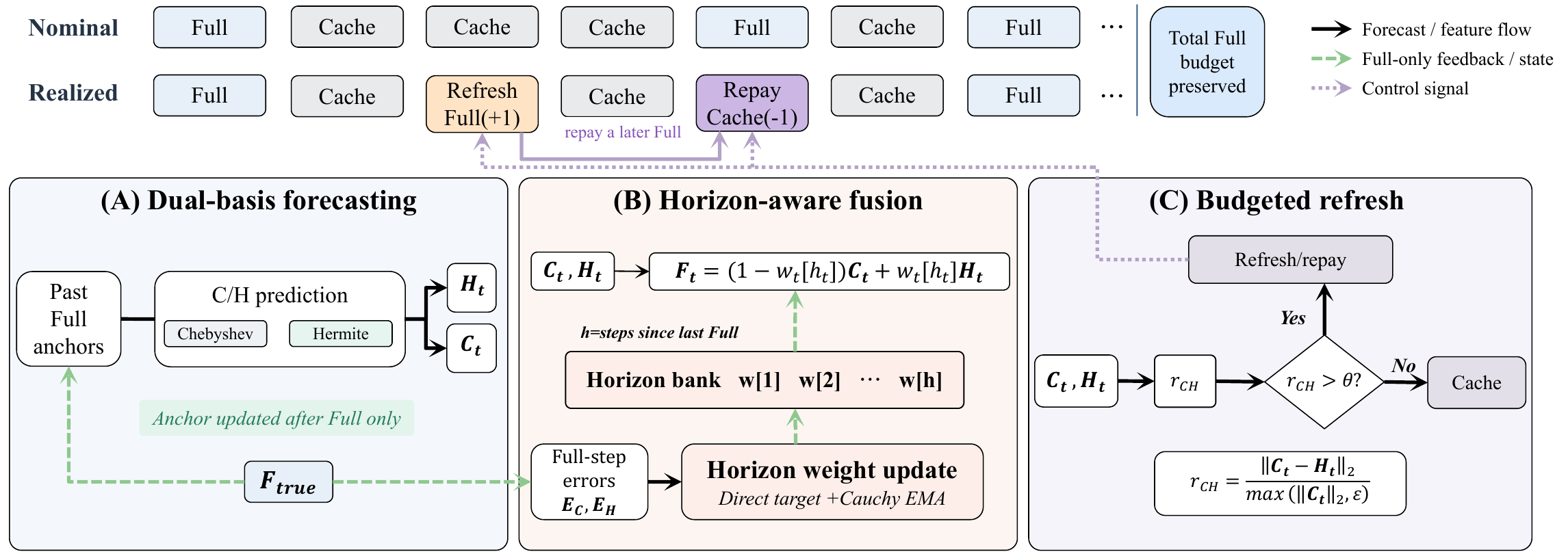}%
 }{%
   \fbox{\parbox[c][1.05in][c]{0.94\textwidth}{\centering
   \textbf{Figure 2 placeholder: Method overview}\\[3pt]
   Historical Full features $\rightarrow$ Chebyshev/Hermite forecasts
   $\rightarrow$ horizon-conditioned fusion $\rightarrow$ Full/Cache;\\
   Full-step prediction errors $\rightarrow$ reliability feedback
   $\rightarrow$ horizon weight bank.
   }}%
 }
 \caption{\textbf{Overview of the proposed feedback-guided feature forecasting framework.} Both predictors use historical features from actual Full evaluations. Cache steps use a horizon-conditioned mixture; eligible Full steps compare their pre-update forecasts with the newly computed feature and update weights for subsequent steps. Optional budget-preserving refresh decisions are separated from the prediction and feedback path.}
 \label{fig:ch-overview}
\end{figure*}

\noindent\textbf{Problem formulation.}
Let $\mathcal S_0\subseteq\{0,\ldots,T-1\}$ denote a prescribed set of Full steps and $N=|\mathcal S_0|$ its evaluation budget. Denote the realized Full-step set by $\mathcal S$; without schedule adaptation, $\mathcal S=\mathcal S_0$. Our primary goal is to improve the feature prediction $\widehat{\bm{F}}_t^{(L)}$ at Cache steps while keeping $|\mathcal S|=N$. This amounts to reducing
\begin{equation}
 \mathbb E\!\left[\sum_{t\notin\mathcal S}
 \bigl\|\widehat{\bm{F}}_t^{(L)}-\bm{F}_t^{(L)}\bigr\|_2^2\right],
 \qquad |\mathcal S|=N.
 \label{eq:ch-objective}
\end{equation}
Here, $\bm{F}_t^{(L)}$ at a Cache step denotes the counterfactual feature that would be obtained by fully evaluating the denoiser at the same accelerated latent $\bm{x}_t$. Because this feature is unavailable without an additional Full evaluation, the objective cannot be measured directly at Cache steps. We instead estimate predictor reliability online from Full evaluations that are already part of the computation budget. As illustrated in Fig.~\ref{fig:ch-overview}, the proposed framework combines Chebyshev and Hermite feature forecasters with Full-step error feedback and horizon-aware fusion. The budget-preserving refresh mechanism is described separately in Sec.~\ref{sec:budgeted_refresh}.

\subsection{Dual-Basis Feature Forecasting}
\label{sec:dual_basis}
We use two feature forecasters with different approximation rules. Chebyshev regression fits historical Full-step features, while Hermite extrapolation uses recursive finite differences.  

Specifically, we adopt the ridge-regularized Chebyshev forecaster of Spectrum~\cite{han2026adaptive} and adapt the scaled-Hermite construction of HiCache~\cite{feng2026hicache}. For Chebyshev forecasting, we use normalized log-SNR as the temporal coordinate. Let $\tau_t\in[-1,1]$ denote this coordinate, and define $\bm{X}_{ik}=\mathsf T_k(\tau_{a_i})$ for polynomial degrees $k=0,\ldots,M$. By stacking the retained anchor features into $\bm{Y}^{(\ell)}$, we obtain
\begin{equation}
 \begin{aligned}
  \bm{B}^{(\ell)}&=(\bm{X}^\top \bm{X}+\lambda_C \bm{I})^{-1}\bm{X}^\top \bm{Y}^{(\ell)},\\
  \bm{C}_t^{(\ell)}&=\sum_{k=0}^{M}\mathsf T_k(\tau_t)\bm{B}_k^{(\ell)}.
 \end{aligned}
 \label{eq:ch-chebyshev}
\end{equation}
Here, $M$ is the maximum Chebyshev polynomial degree, $\mathsf T_k$ denotes the Chebyshev polynomial of the first kind~\cite{mason2002chebyshev} of degree $k$, and $\lambda_C$ is the ridge-regularization coefficient. The matrix $\bm B^{(\ell)}$ contains the fitted polynomial coefficients, with $\bm B_k^{(\ell)}$ denoting the coefficients of degree $k$, and $\bm{I}$ is the identity matrix. The temporal coordinate $\tau_t$ is determined solely by the known noise schedule and does not require future feature observations.

For Hermite forecasting, we adopt the scaled basis of HiCache and update recursive finite differences using the actual gaps between Full anchors. At each actual Full anchor, recursive finite differences are updated by
\begin{equation}
 \begin{aligned}
  \bm{D}_{j,0}^{(\ell)}&=\bm{F}_{a_j}^{(\ell)},\\
  \bm{D}_{j,k}^{(\ell)}&=
  \frac{\bm{D}_{j,k-1}^{(\ell)}-\bm{D}_{j-1,k-1}^{(\ell)}}{a_j-a_{j-1}}.
 \end{aligned}
 \label{eq:ch-differences}
\end{equation}
Once the required history is available, the order-$P$ Hermite forecast is
\begin{equation}
 \bm{H}_t^{(\ell)}=\bm{D}_{j,0}^{(\ell)}+
 \sum_{k=1}^{P}\frac{s^k}{k!}\mathsf H_k(s h_t)\bm{D}_{j,k}^{(\ell)},
 \label{eq:ch-hermite}
\end{equation}
where $s$ is a scale and $\mathsf H_k$ follows the physicists' convention~\cite{abramowitz1964handbook}: $\mathsf H_0(u)=1$, $\mathsf H_1(u)=2u$, and $\mathsf H_k(u)=2u\mathsf H_{k-1}(u)-2(k-1)\mathsf H_{k-2}(u)$. The recurrence uses the actual, potentially unequal Full gaps; it is a finite-difference forecasting rule rather than exact nonuniform polynomial interpolation.

\subsection{Horizon-Aware Forecast Fusion}
\label{sec:horizon_aware}

\noindent\textbf{Horizon-indexed mixture.}
Because predictor reliability varies with the distance from the most recent Full anchor, we maintain a horizon-indexed Hermite weight $w_t[h]\in[0,1]$, initialized to $w_0$ for each sample. At a Cache step, the fused forecast is
\begin{equation}
 \widehat{\bm{F}}_t^{(L)}=(1-w_t[h_t])\bm{C}_t^{(L)}+w_t[h_t]\bm{H}_t^{(L)}.
 \label{eq:ch-mixture}
\end{equation}
This formulation mixes the two forecasts, with the weight adapted online through the reliability feedback described below.

\noindent\textbf{Full-step reliability feedback.}
At each eligible Full step, both forecasts are computed from preceding Full anchors before the current observation is added to the feature history. We compare these forecasts with the newly computed feature to obtain their prediction errors:
\begin{equation}
 E_{i,t}=\frac{\|\bm{P}_{i,t}^{(L)}-\bm{F}_t^{(L)}\|_2^2}
                    {\|\bm{F}_t^{(L)}\|_2^2+\epsilon_e},\quad i\in\{C,H\},
 \label{eq:ch-errors}
\end{equation}
where $\bm{P}_{C,t}^{(L)}=\bm{C}_t^{(L)}$ and $\bm{P}_{H,t}^{(L)}=\bm{H}_t^{(L)}$. Here, $\epsilon_e>0$ is a small numerical stabilization constant. Errors and mixture states are computed per sample, including separate branches when classifier-free guidance is used. We then define the Hermite reliability target using normalized inverse-error weighting:
\begin{equation}
 v_t=\frac{E_{C,t}+\epsilon_e}
             {E_{C,t}+E_{H,t}+2\epsilon_e}.
 \label{eq:ch-target}
\end{equation}
A smaller Hermite error relative to the Chebyshev error yields a higher target weight.

\noindent\textbf{Cross-horizon transfer.}
Full evaluations provide reliability feedback at only a few horizons. We transfer each observation to nearby horizons using a Cauchy-shaped kernel:
\begin{equation}
 \begin{aligned}
  \eta_t(h)&=\frac{\beta}{1+((h-h_t)/r)^2},\\
  w_{t+1}[h]&=(1-\eta_t(h))w_t[h]+\eta_t(h)v_t.
 \end{aligned}
 \label{eq:ch-update}
\end{equation}
Here, $\beta\in[0,1]$ controls feedback strength, and $r>0$ determines how quickly feedback decays with horizon distance. At the observed horizon $h=h_t$, the update rate equals $\beta$. For other horizons, it decreases with the distance $|h-h_t|$, allowing each Full-step observation to update several horizon weights while assigning greater importance to nearby horizons. The convex update preserves $w_t[h]\in[0,1]$. Cache steps leave the bank unchanged, and refreshing feature anchors does not reset it.

We define the normalized separation between the two forecasts as
\begin{equation}
 I_t=\frac{\|\bm{H}_t^{(L)}-\bm{C}_t^{(L)}\|_2}
               {\|\bm{F}_t^{(L)}\|_2+\epsilon_e}.
 \label{eq:ch-impact}
\end{equation}
Feedback is discarded when $I_t<\delta$, where $\delta>0$ is a minimum-separation threshold, or when either prediction is missing or nonfinite. Warmup also leaves reliability unchanged. The initial weight is used until the configured minimum number of valid observations is reached. Only already-decided Full steps provide supervision, and their feedback affects subsequent iterations.

\subsection{Budgeted Refresh}
\label{sec:budgeted_refresh}

\noindent\textbf{Disagreement observer.}
Adaptive fusion does not eliminate forecasting errors, and both forecasters may still become unreliable. To guide cache refreshing, we measure their normalized disagreement at an intermediate signal layer $p$:
\begin{equation}
 r_t^{(p)}=\frac{\|\bm{C}_t^{(p)}-\bm{H}_t^{(p)}\|_2}
                    {\max(\|\bm{C}_t^{(p)}\|_2,\epsilon_g)}.
 \label{eq:ch-disagreement}
\end{equation}
Here, $\epsilon_g>0$ is a numerical stabilization constant that prevents division by zero. Both forecasts are derived from the Full-step history of layer $p$; evaluating this signal at a Cache step requires no additional Transformer-block computation. The signal layer $p$ need not equal the cached-output layer $L$, whose fused forecast is given by Eq.~\eqref{eq:ch-mixture}. Predictor disagreement is only a proxy for forecasting uncertainty, not an error bound, because the two predictors can make correlated errors.

\noindent\textbf{Refresh and repay.}
For budget-preserving schedule adaptation, we follow RACER's refresh-and-repay rule~\cite{li2026disagree}. Starting from the nominal Full-step set $\mathcal S_0$, let $d_t$ denote the number of outstanding refreshes that must be repaid, and define the number of remaining nominal Full slots as
\begin{equation}
 m_t=|\{s\in\mathcal S_0:s>t\}|.
 \label{eq:ch-slots}
\end{equation}
After warmup, a nominal Cache step is promoted to a Full step when $r_t^{(p)}>\theta_{p,N}$ and $d_t<m_t$; the promotion increases $d_t$ by one. Here, $\theta_{p,N}$ denotes the disagreement threshold associated with signal layer $p$ and Full-evaluation budget $N$. Conversely, when a later nominal Full step is encountered with outstanding debt, it is replaced by a Cache step, reducing $d_t$ by one. All other nominal actions are unchanged, and the nominal schedule advances independently of these substitutions. Provided that every remaining post-warmup nominal Full slot is eligible for repayment, the total number of Full evaluations is preserved:
\begin{equation}
 \begin{aligned}
  N_{\mathrm{actual}}&=N+N_{\mathrm{refresh}}-N_{\mathrm{repay}}\\
                    &=N.
 \end{aligned}
 \label{eq:ch-budget}
\end{equation}
The condition $d_t<m_t$ ensures that each accepted refresh can be matched to an available future repayment slot; consequently, no refresh debt remains at the end of sampling.
\section{Experiments}
\label{sec:experiments}
\begin{table*}[t]
\centering
\small
\setlength{\tabcolsep}{10pt}
\caption{Quantitative comparison of text-to-image generation task on 200 DrawBench prompts with FLUX.1-dev. Bold denotes the best image-quality metric at each NFE budget.}
\label{tab:flux}
\resizebox{\textwidth}{!}{%
\begin{tabular}{rlrrrrrrrr}
\toprule
NFE & Method & PSNR$\uparrow$ & SSIM$\uparrow$ & LPIPS$\downarrow$ & IR$\uparrow$ & CLIP$\uparrow$ & FLOPs (T)$\downarrow$ & Latency (s)$\downarrow$ & Speedup$\uparrow$ \\
\midrule
50 & Reference & -- & -- & -- & 0.9986 & 27.224 & 3734.56 & 40.43 & $1.00\times$ \\
\midrule
18 & TaylorSeer & 19.52 & 0.724 & 0.237 & \textbf{1.007} & \textbf{27.52} & 1354.98 & 15.73 & $2.76\times$ \\
18 & HiCache & 19.90 & 0.736 & 0.229 & 0.979 & 27.36 & 1354.98 & 15.90 & $2.76\times$ \\
18 & Spectrum & 24.35 & 0.848 & 0.141 & 0.997 & 27.43 & 1354.80 & 14.84 & $2.76\times$ \\
18 & RACER & 24.51 & 0.849 & 0.144 & 0.972 & 27.29 & 1354.80 & 14.92 & $2.76\times$ \\
\rowcolor{oursgray}
18 & CH-Cache & \textbf{25.66} & \textbf{0.866} & \textbf{0.122} & 0.990 & 27.38 & 1355.07 & 15.24 & $2.76\times$ \\
\midrule
14 & TaylorSeer & 17.71 & 0.668 & 0.307 & \textbf{0.999} & 27.29 & 1057.53 & 12.62 & $3.53\times$ \\
14 & HiCache & 18.38 & 0.685 & 0.286 & 0.997 & \textbf{27.46} & 1057.53 & 12.82 & $3.53\times$ \\
14 & Spectrum & 24.21 & 0.846 & 0.147 & 0.985 & 27.25 & 1057.32 & 11.61 & $3.53\times$ \\
14 & RACER & 23.99 & 0.836 & 0.167 & 0.971 & 27.27 & 1057.32 & 11.67 & $3.53\times$ \\
\rowcolor{oursgray}
14 & CH-Cache & \textbf{24.89} & \textbf{0.851} & \textbf{0.142} & 0.979 & 27.34 & 1057.53 & 11.92 & $3.53\times$ \\
\midrule
10 & TaylorSeer & 15.71 & 0.598 & 0.408 & 0.950 & 27.37 & 760.08 & 9.50 & $4.91\times$ \\
10 & HiCache & 16.85 & 0.627 & 0.364 & \textbf{0.985} & \textbf{27.59} & 760.08 & 9.83 & $4.91\times$ \\
10 & Spectrum & 22.12 & 0.777 & 0.236 & 0.984 & 27.39 & 759.84 & 8.34 & $4.91\times$ \\
10 & RACER & 22.05 & 0.771 & 0.254 & 0.953 & 27.43 & 759.84 & 8.42 & $4.91\times$ \\
\rowcolor{oursgray}
10 & CH-Cache & \textbf{22.62} & \textbf{0.790} & \textbf{0.224} & 0.982 & 27.44 & 759.99 & 8.66 & $4.91\times$ \\
\midrule
9 & TaylorSeer & 15.05 & 0.570 & 0.457 & 0.887 & 27.33 & 685.72 & 8.73 & $5.45\times$ \\
9 & HiCache & 16.35 & 0.607 & 0.397 & 0.962 & 27.69 & 685.72 & 9.15 & $5.45\times$ \\
9 & Spectrum & 20.77 & 0.717 & 0.339 & 0.962 & \textbf{27.73} & 685.47 & 7.57 & $5.45\times$ \\
9 & RACER & 21.39 & 0.740 & 0.301 & 0.945 & 27.55 & 685.47 & 7.64 & $5.45\times$ \\
\rowcolor{oursgray}
9 & CH-Cache & \textbf{21.88} & \textbf{0.766} & \textbf{0.256} & \textbf{0.964} & 27.53 & 685.61 & 7.85 & $5.45\times$ \\
\midrule
8 & TaylorSeer & 14.38 & 0.547 & 0.490 & 0.837 & 27.06 & 611.36 & 8.01 & $6.11\times$ \\
8 & HiCache & 15.80 & 0.583 & 0.430 & \textbf{0.949} & 27.59 & 611.36 & 8.42 & $6.11\times$ \\
8 & Spectrum & 20.03 & 0.661 & 0.407 & 0.868 & 27.65 & 611.10 & 6.77 & $6.11\times$ \\
8 & RACER & \textbf{20.45} & 0.685 & 0.371 & 0.854 & 27.62 & 611.10 & 6.83 & $6.11\times$ \\
\rowcolor{oursgray}
8 & CH-Cache & 20.22 & \textbf{0.689} & \textbf{0.364} & 0.882 & \textbf{27.71} & 611.23 & 7.01 & $6.11\times$ \\
\bottomrule
\end{tabular}
}
\end{table*}
\begin{table*}[t]
\centering
\small
\setlength{\tabcolsep}{10pt}
\caption{Quantitative comparison of text-to-image generation task on 200 DrawBench prompts with SD3.5-Large. Bold denotes the best image-quality metric at each NFE budget.}
\label{tab:sd35}
\resizebox{\textwidth}{!}{%
\begin{tabular}{rlrrrrrrrr}
\toprule
NFE & Method & PSNR$\uparrow$ & SSIM$\uparrow$ & LPIPS$\downarrow$ & IR$\uparrow$ & CLIP$\uparrow$ & FLOPs (T)$\downarrow$ & Latency (s)$\downarrow$ & Speedup$\uparrow$ \\
\midrule
50 & Reference & -- & -- & -- & 1.0698 & 28.609 & 3127.60 & 36.20 & $1.00\times$ \\
\midrule
18 & TaylorSeer & 11.475 & 0.5386 & 0.5084 & 0.7593 & 28.037 & 1136.78 & 16.71 & $2.75\times$ \\
18 & HiCache & 12.275 & 0.5723 & 0.4655 & 0.9073 & 28.272 & 1136.78 & 16.16 & $2.75\times$ \\
18 & Spectrum & 18.361 & 0.7590 & 0.2319 & 1.0118 & 28.544 & 1136.48 & 13.44 & $2.75\times$ \\
\rowcolor{oursgray}
18 & CH-Cache & \textbf{18.663} & \textbf{0.7713} & \textbf{0.2181} & \textbf{1.0322} & \textbf{28.642} & 1136.91 & 13.98 & $2.75\times$ \\
\midrule
14 & TaylorSeer & 10.266 & 0.4703 & 0.6007 & 0.4709 & 25.472 & 887.93 & 14.13 & $3.52\times$ \\
14 & HiCache & 10.909 & 0.5024 & 0.5514 & 0.7127 & 27.977 & 887.93 & 13.70 & $3.52\times$ \\
14 & Spectrum & \textbf{17.883} & 0.7363 & 0.2669 & 0.9686 & 28.496 & 887.57 & 10.49 & $3.52\times$ \\
\rowcolor{oursgray}
14 & CH-Cache & 17.858 & \textbf{0.7390} & \textbf{0.2558} & \textbf{1.0185} & \textbf{28.600} & 887.91 & 11.00 & $3.52\times$ \\
\midrule
10 & TaylorSeer & 9.323 & 0.4084 & 0.7093 & -0.1408 & 24.818 & 639.08 & 11.80 & $4.89\times$ \\
10 & HiCache & 9.694 & 0.4202 & 0.6759 & 0.1179 & 26.420 & 639.08 & 12.02 & $4.89\times$ \\
10 & Spectrum & 15.656 & 0.6089 & 0.3967 & 0.8018 & 27.890 & 638.67 & 7.54 & $4.90\times$ \\
\rowcolor{oursgray}
10 & CH-Cache & \textbf{15.757} & \textbf{0.6369} & \textbf{0.3566} & \textbf{0.8987} & \textbf{28.338} & 638.93 & 8.02 & $4.90\times$ \\
\midrule
9 & TaylorSeer & 9.172 & 0.3960 & 0.7337 & -0.4139 & 25.118 & 576.86 & 11.20 & $5.42\times$ \\
9 & HiCache & 9.431 & 0.4026 & 0.7109 & -0.1956 & 25.328 & 576.86 & 11.54 & $5.42\times$ \\
9 & Spectrum & \textbf{14.998} & 0.5482 & 0.4700 & 0.4667 & 26.812 & 576.44 & 6.83 & $5.43\times$ \\
\rowcolor{oursgray}
9 & CH-Cache & 14.745 & \textbf{0.6042} & \textbf{0.4095} & \textbf{0.7887} & \textbf{28.019} & 576.68 & 7.31 & $5.42\times$ \\
\midrule
8 & TaylorSeer & 9.074 & 0.3807 & 0.7522 & -0.5702 & 24.595 & 514.65 & 10.58 & $6.08\times$ \\
8 & HiCache & 9.209 & 0.3822 & 0.7405 & -0.4959 & 24.261 & 514.65 & 10.81 & $6.08\times$ \\
8 & Spectrum & \textbf{14.285} & 0.4894 & 0.4902 & 0.3447 & 26.907 & 514.22 & 6.10 & $6.08\times$ \\
\rowcolor{oursgray}
8 & CH-Cache & 13.754 & \textbf{0.5507} & \textbf{0.4660} & \textbf{0.5718} & \textbf{27.776} & 514.44 & 6.53 & $6.08\times$ \\
\bottomrule
\end{tabular}
}
\end{table*}

\subsection{Experimental Setup}
\label{sec:exp_setup}

\noindent\textbf{Models and evaluation.}
We evaluate our method on two text-to-image diffusion transformers, FLUX.1-dev~\cite{flux2024} and SD3.5-Large~\cite{esser2024scaling}. We report five budgets of 18, 14, 10, 9, and 8 full denoiser evaluations over a 50-step trajectory. All experiments use a fixed random seed of 42. The FLUX and SD3.5 configurations use guidance scales of 3.5 and 7.0, respectively. PSNR, SSIM~\cite{wang2004image}, and LPIPS~\cite{zhang2018unreasonable} measure fidelity to the reference images used by the evaluation pipeline, whereas ImageReward (IR)~\cite{xu2023imagereward} and CLIPScore~\cite{hessel2021clipscore} evaluate image--text preference and alignment. We report FLOPs, measured latency, and FLOPs-based speedup, defined as the ratio of full-reference FLOPs to accelerated-method FLOPs.

\noindent\textbf{Compared methods.}
We compare CH-Cache with representative feature forecasting methods, including TaylorSeer~\cite{liu2025reusing}, HiCache~\cite{feng2026hicache}, Spectrum~\cite{han2026adaptive}, and RACER~\cite{li2026disagree}. All baselines use their published configurations.

\noindent\textbf{Implementation details.}
CH-Cache uses Chebyshev and Hermite forecasts, error feedback with a horizon-dependent weight bank, and selected-layer C/H disagreement. We retain five warmup steps and the inherited refresh-and-repay budget control~\cite{li2026disagree}. The signal is read at global block 48 for FLUX and block 29 for SD3.5.

\subsection{Main Results}
\noindent\textbf{FLUX.1-dev.}
Table~\ref{tab:flux} compares the methods under five target NFE budgets. For NFE budgets of $18$, $14$, $10$, $9$, and $8$, respectively, we initialize the Chebyshev/Hermite weights $(w_C^0,w_H^0)$ to $(0.2,0.8)$, $(0.2,0.8)$, $(0.5,0.5)$, $(0.2,0.8)$, and $(0.8,0.2)$, with corresponding feedback rates $\beta$ of $0.2$, $0.2$, $0.5$, $0.2$, and $0.5$. These budget-specific configurations are selected on a separate development set by maximizing a composite score of PSNR, ImageReward, and CLIPScore among the evaluated budget-matched candidates. This criterion balances reference fidelity, image--text preference, and semantic alignment, rather than optimizing a single metric. Our method achieves the best SSIM and LPIPS in all five groups and the highest PSNR in four of them. At NFE$=10$, it improves over RACER from 22.050 to 22.621 PSNR and from 0.254 to 0.224 LPIPS, while increasing IR from 0.953 to 0.982. Relative to Spectrum at the same budget, PSNR increases by 0.498 dB and LPIPS decreases by 0.012. These results show that the adaptive C/H mixture can improve fidelity to the 50-step reference under substantial feature reuse. The gains are not uniform across metrics: at NFE$=8$, RACER obtains a slightly higher PSNR (20.45 versus 20.22 dB), while our method retains better SSIM and LPIPS. ImageReward and CLIPScore also exhibit method-dependent trade-offs rather than consistent improvements at every budget.

\noindent\textbf{SD3.5-Large.}
Table~\ref{tab:sd35} reports the SD3.5-Large results across five NFE budgets. For NFE budgets of $18$, $14$, $10$, $9$, and $8$, respectively, the initial Chebyshev/Hermite weights $(w_C^0,w_H^0)$ are $(0.7,0.3)$, $(0.7,0.3)$, $(0.3,0.7)$, $(0.7,0.3)$, and $(0.8,0.2)$, and the corresponding feedback rates $\beta$ are $0.2$, $0.5$, $0.5$, $0.5$, and $0.2$. These configurations are selected using the same development-set composite criterion as for FLUX. Among these reported methods, our method achieves the highest SSIM, lowest LPIPS, highest ImageReward, and highest CLIPScore at all five NFE budgets. For instance, at NFE$=10$, the Spectrum obtains 0.6089 SSIM and 0.3967 LPIPS, whereas our method achieves 0.6369 SSIM and 0.3566 LPIPS. Its ImageReward and CLIPScore also increase from 0.8018 and 27.890 to 0.8987 and 28.338, respectively. PSNR is less uniform: our method leads at NFE$=18$ and $10$, but Spectrum achieves slightly higher PSNR at NFE$=14$, $9$, and $8$. These observations show improvements in structural and perceptual fidelity for the displayed SD3.5-Large baselines while also indicating that the benefit depends on the evaluation metric and operating point.

\subsection{Ablation Studies}
\label{sec:ablation}
We study output fusion, refresh control, and signal location on FLUX.1-dev using 200 DrawBench~\cite{saharia2022photorealistic} prompts. Within each comparison, we keep the sampling configuration and total number of Full evaluations fixed. These experiments evaluate closed-loop generation: changing fusion can also change subsequent features and refresh decisions.

\begin{figure}[t]
    \centering
    \includegraphics[width=\columnwidth]
        {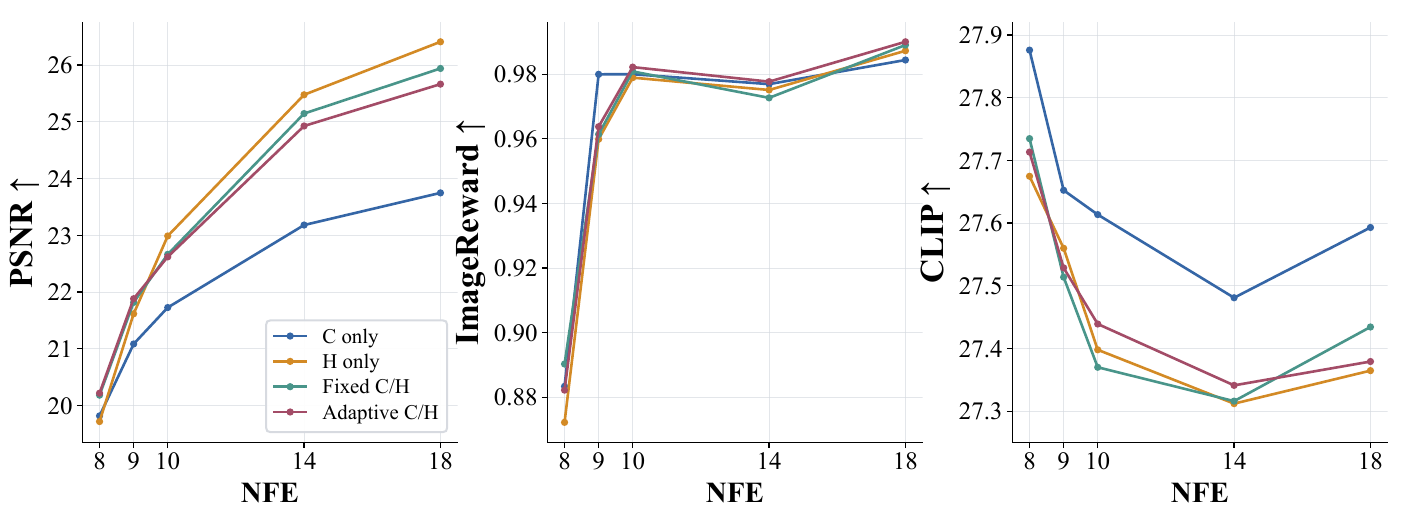}
    \caption{\textbf{Effect of dual-basis feature forecasting.}
    Comparison of Chebyshev-only, Hermite-only,
    fixed C/H fusion, and adaptive C/H fusion
    on FLUX.1-dev across five NFE budgets.
    PSNR, ImageReward, and CLIPScore are reported.
    All variants retain the C/H disagreement-based
    refresh observer.}
    \label{fig:ab-fusion}
\end{figure}

\noindent\textbf{Effect of dual-basis feature forecasting.}
We investigate the contributions of Chebyshev and Hermite forecasting by comparing Chebyshev-only (C), Hermite-only (H), fixed C/H fusion, and adaptive C/H fusion. The single-output controls retain both predictors for disagreement-based gating; they isolate output fusion rather than removing a predictor from the controller. As shown in Fig.~\ref{fig:ab-fusion}, adaptive fusion achieves the highest PSNR at NFE$=8$ and $9$, improving over fixed fusion by 0.035 and 0.065 dB, respectively. This suggests that combining the two predictors is particularly beneficial under aggressive caching. At higher NFE budgets, Hermite-only forecasting achieves better reconstruction fidelity, indicating that the benefits of fusion are budget-dependent. Meanwhile, Chebyshev-only forecasting generally obtains higher CLIP scores, while ImageReward exhibits relatively small differences among the variants. These results highlight the complementary characteristics of the two forecasters.

\begin{figure}[t]
    \centering
    \includegraphics[width=\columnwidth]{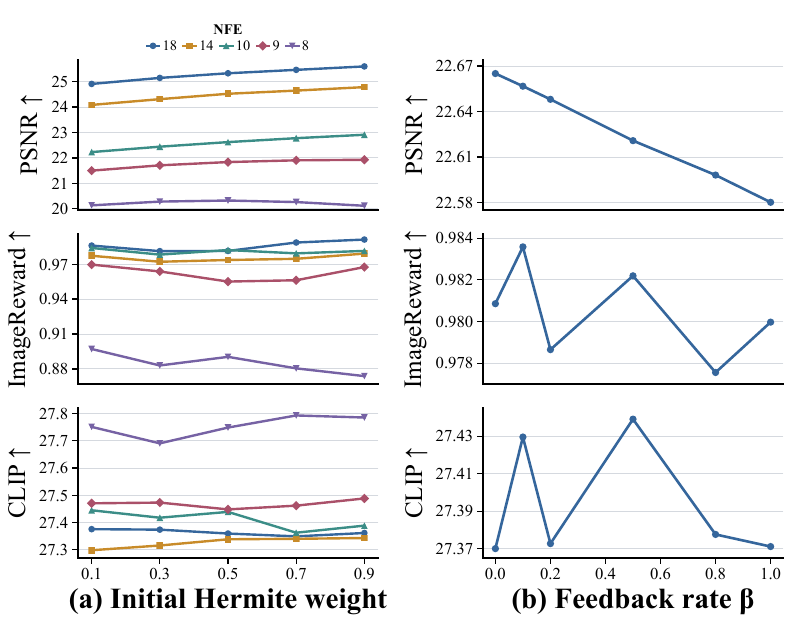}
    \caption{\textbf{Hyperparameter sensitivity on FLUX.1-dev.}
    \textbf{(a)} Sensitivity to the initial Hermite weight across five NFE budgets, with the feedback rate fixed at $\beta=0.5$.
    \textbf{(b)} Sensitivity to the feedback rate at NFE$=10$, with the initial Hermite weight fixed at $0.5$.
    All results are averaged over 200 DrawBench prompts.}
    \label{fig:param-ablation}
\end{figure}

\noindent\textbf{Sensitivity to initialization and feedback rate.}
We investigate the effects of the initial Hermite weight $w_0$ and feedback rate $\beta$ on FLUX.1-dev. As shown in Fig.~\ref{fig:param-ablation}(a), increasing $w_0$ generally improves PSNR at NFE$=9$, $10$, $14$, and $18$, whereas NFE$=8$ favors a more balanced fusion. ImageReward and CLIPScore exhibit different preferences, indicating that the optimal initialization depends on both the computation budget and evaluation metric. Fig.~\ref{fig:param-ablation}(b) shows that PSNR gradually decreases with increasing $\beta$ at NFE$=10$, while ImageReward and CLIPScore peak at $\beta=0.1$ and $\beta=0.5$, respectively. At $\beta=0$, the fusion weights remain fixed at their initial values, providing a fixed-weight reference within the same setting. This configuration yields the highest PSNR, whereas nonzero feedback rates favor ImageReward or CLIPScore. These results suggest that moderate feedback can benefit semantic metrics, although stronger adaptation does not necessarily improve reconstruction fidelity.

\noindent\textbf{Refresh and repay.}
We disable refresh at NFE$=10$ while keeping the feedback fusion unchanged. This yields the nominal Full schedule $[0,1,2,3,4,6,11,19,30,44]$ (zero-based); enabling refresh-and-repay changes the schedule for 94 of 200 prompts, with an average of 1.06 refreshes, while preserving ten Full evaluations. As shown in Table~\ref{tab:ab-refresh}, enabling the refresh-and-repay mechanism yields modest improvements in the reported metrics. Specifically, PSNR increases from 22.5882 to 22.6208 dB, while ImageReward and CLIPScore also exhibit slight gains. These results suggest that adaptively adjusting the placement of Full evaluations can provide incremental benefits without increasing the overall computation budget. Although the improvements are relatively small, they indicate the potential of disagreement-guided refresh as a complementary component to adaptive feature forecasting.

\begin{table}[!t]
\centering
\small
\caption{Refresh ablation on FLUX at NFE$=10$ with unchanged C/H feedback fusion. Both configurations use 200 DrawBench prompts and the same total Full budget.}
\label{tab:ab-refresh}
\begin{tabular*}{\columnwidth}{@{\extracolsep{\fill}}lcccc@{}}
\toprule
Refresh & NFE & PSNR$\uparrow$ & IR$\uparrow$ & CLIP$\uparrow$ \\
\midrule
Off & 10 & 22.5882 & 0.98206 & 27.4328 \\
On  & 10 & \textbf{22.6208} & \textbf{0.98219} & \textbf{27.4391} \\
\bottomrule
\end{tabular*}
\end{table}

\noindent\textbf{Location of the disagreement signal.}
We evaluate global blocks 10, 19, 20, 29, 39, 48, and 57 at NFE$=10$, using image-token features in both the joint and single-stream stages. Each layer receives its own Q90 threshold, estimated on the same refresh-disabled development trajectories, to account for disagreement-scale differences. The cached-output layer remains fixed, while the disagreement signal layer and its calibrated threshold vary across configurations. \Cref{tab:ab-layers} shows a metric-dependent trade-off: block 57 gives the highest PSNR (22.553 dB), block 10 the highest ImageReward (0.98544), and block 48 the highest CLIP score (27.626). Compared with block 57, block 48 improves CLIPScore from 27.555 to 27.626, with a PSNR decrease of only 0.026 dB. Relative to block 10, it achieves a 0.211 higher CLIPScore but a 0.0075 lower ImageReward. We therefore select block 48 as a semantic-alignment-oriented operating point rather than assuming that one layer is optimal for all metrics.

\begin{table}[!t]
\centering\small
\caption{Signal-layer ablation on FLUX.1-dev at NFE$=10$, averaged over 200 DrawBench prompts. Each layer uses an independently calibrated Q90 threshold. Global block indices are 1-based: 1--19 are joint blocks and 20--57 are single-stream blocks.}
\label{tab:ab-layers}
\begin{tabular*}{\columnwidth}{@{\extracolsep{\fill}}rrrr@{}}
\toprule
Layer & PSNR $\uparrow$ & IR $\uparrow$ & CLIP $\uparrow$ \\
\midrule
10 & 22.523 & \textbf{0.9854} & 27.415 \\
19 & 22.421 & 0.9705 & 27.400 \\
20 & 22.431 & 0.9722 & 27.419 \\
29 & 22.454 & 0.9729 & 27.485 \\
39 & 22.546 & 0.9761 & 27.594 \\
48 & 22.527 & 0.9779 & \textbf{27.626} \\
57 & \textbf{22.553} & 0.9790 & 27.555 \\
\bottomrule
\end{tabular*}
\end{table}

\section{Conclusion}
\label{sec:conclusion}

We presented CH-Cache, a training-free framework for diffusion acceleration that adaptively fuses Chebyshev and Hermite feature forecasts through online reliability feedback. By leveraging prediction errors observed at existing Full evaluations, CH-Cache updates horizon-indexed fusion weights without additional denoiser evaluations. Predictor disagreement further guides budget-preserving cache refreshing through RACER's refresh-and-repay mechanism. Experiments on FLUX.1-dev and SD3.5-Large demonstrate improved reference fidelity across multiple matched computation budgets, while ablation studies highlight the benefits of complementary forecasting and online adaptation. Nevertheless, the effectiveness of adaptive weighting and refresh control varies across budgets and evaluation metrics, revealing trade-offs between reconstruction fidelity and text-image alignment. Future work will explore more robust cross-horizon feedback propagation and quality-aware fusion strategies across different sampling budgets.

\section*{LLM Usage Statement}
We used ChatGPT (OpenAI) to assist with manuscript drafting, language refinement, related-work exploration, and the organization of experimental analysis. OpenAI Codex was used to assist with code implementation and debugging. These tools served as research and writing assistants, while the authors remained responsible for the research methodology, experimental validation, technical claims, and final manuscript.

{
    \small
    \bibliographystyle{ieeenat_fullname}
    \bibliography{main}
}

% WARNING: do not forget to delete the supplementary pages from your submission 
\clearpage
\setcounter{page}{1}
\maketitlesupplementary

\appendix

\section{Implementation Details}
\label{sec:implementation}

\subsection{Detailed Algorithm}
\label{sec:supp_algorithm}

Algorithm~\ref{alg:ch_cache} summarizes the inference procedure of CH-Cache. All forecasts are computed from preceding Full anchors. Valid reliability feedback is obtained before updating the predictor history, while Cache steps leave both the history and weight bank unchanged.

\begin{algorithm}[H]
\caption{Inference with CH-Cache}
\label{alg:ch_cache}
\begin{algorithmic}[1]

\Require Initial latent $x_0$, nominal schedule
$\mathcal{S}_0$, sampling steps $T$.
\Ensure Final latent $x_T$.

\State Initialize $\mathcal{A}\gets\emptyset$,
$w[h]\gets w_0$, $d,n\gets0$, $a\gets-1$

\For{$t=0,\ldots,T-1$}
    \State $h_t\gets t-a$,
    $m_t\gets|\{s\in\mathcal{S}_0:s>t\}|$
    \State $q_t\gets\mathbb{I}[t\in\mathcal{S}_0]$

    \If{$t\geq W \land q_t=1 \land d>0$}
        \State $q_t\gets0$, $d\gets d-1$
        \Comment{Repay}
    \ElsIf{$t\geq W \land q_t=0 \land d<m_t$}
        \State $r_t^{(p)}\gets
        \Call{Disagreement}{\mathcal{A},p,t}$
        \If{$r_t^{(p)}>\theta_{p,N}$}
            \State $q_t\gets1$, $d\gets d+1$
            \Comment{Refresh}
        \EndIf
    \EndIf

    \State $C_t,H_t\gets
    \Call{Forecast}{\mathcal{A},t}$

    \If{$q_t=1$}
        \State $u_t,\mathcal{F}_t\gets
        \Call{FullForward}{x_t,t}$
        \If{\Call{ValidFeedback}{$C_t,H_t,\mathcal{F}_t,t$}}
            \State Update $w$ via Eqs.~(7)--(9);
            $n\gets n+1$
        \EndIf
        \State Update $\mathcal{A}$ with $\mathcal{F}_t$;
        $a\gets t$
    \Else
        \State $\widehat{F}_t\gets
        (1-\tilde{w}_t)C_t+\tilde{w}_t H_t$
        \State $u_t\gets
        \Call{CacheForward}{x_t,t,\widehat{F}_t}$
    \EndIf

    \State $x_{t+1}\gets
    \Call{SchedulerStep}{x_t,u_t,t}$
\EndFor
\State \Return $x_T$

\end{algorithmic}
\end{algorithm}

Here, $d$ denotes the refresh debt, and $n$ counts valid feedback observations. \textsc{Disagreement} computes the signal-layer C/H forecasts from preceding Full anchors and evaluates $r_t^{(p)}$ using Eq.~\eqref{eq:ch-disagreement}. Feedback validity follows Eq.~\eqref{eq:ch-impact} and the warmup requirements. The effective weight $\tilde{w}_t$ equals $w[h_t]$ if $n\geq n_{\min}$, and $w_0$ otherwise.

\subsection{Experimental Configurations}
\label{sec:experimental_configurations}

Table~\ref{tab:baseline_configurations} summarizes the main hyperparameters of the compared methods. For NFE budgets $(18,14,10,9,8)$, the refresh intervals are $I_B=(3,4,6,7,8)$, and the window-growth parameters are $\alpha_B=(0.25,0.75,3.0,5.0,7.0)$, respectively. TaylorSeer and HiCache use three initial full-computation steps. Spectrum and RACER use five initial full-computation steps, an initial cache window of two, a Chebyshev degree of four, and a regularization coefficient of $0.1$.

\begin{table}[!h]
    \centering
    \caption{Main hyperparameters of the compared methods.
    $I$ denotes the refresh interval, $O$ the maximum
    expansion order, $s_H$ the HiCache scale, and
    $\alpha$ the window-growth parameter.}
    \label{tab:baseline_configurations}
    \small
    \setlength{\tabcolsep}{5pt}
    \renewcommand{\arraystretch}{1.15}
    \begin{tabular*}{\columnwidth}{c|c|l}
        \hline
        \textbf{Model} & \textbf{Method} & \textbf{Hyperparameters} \\
        \hline
        FLUX.1-dev
        & TaylorSeer & $I=I_B,\ O=1$ \\
        & HiCache   & $I=I_B,\ O=2,\ s_H=0.5$ \\
        & Spectrum  & $\alpha=\alpha_B$ \\
        & RACER     & $\alpha=\alpha_B$ \\
        \hline
        SD3.5-Large
        & TaylorSeer & $I=I_B,\ O=2$ \\
        & HiCache    & $I=I_B,\ O=2,\ s_H=0.5$ \\
        & Spectrum  & $\alpha=\alpha_B$ \\
        \hline
    \end{tabular*}
\end{table}

\section{Ablation Studies}
\label{sec:supp_ablation}

\subsection{Replacing Hermite with Taylor Prediction}
\label{sec:ablation_taylor}

We investigate replacing the Hermite predictor in CH-Cache with a Taylor predictor. Both forecasters use second-order predictions to ensure a fair comparison. We compare C/H and C/T fusion with and without adaptive refresh on FLUX.1-dev and SD3.5-Large. All configurations are evaluated on 200 DrawBench prompts at NFE$=10$ using 50 denoising steps.

For each model, the C/T variant inherits the initial weights, weight-update coefficient, signal layer, and refresh threshold of the corresponding C/H configuration. Online fusion-weight updates remain active when refresh is disabled. The refreshed C/T rows use C/T disagreement. Thus, the refresh-disabled comparison isolates the predictor replacement, whereas the refresh-enabled comparison also changes the disagreement signal.

\begin{table}[t]
    \centering
    \caption{Hermite versus Taylor prediction
    at NFE$=10$. ``None'' disables adaptive refresh;
    C/H and C/T indicate the disagreement used to trigger
    refresh. Best results within each model are bold.}
    \label{tab:taylor_replacement}
    \footnotesize
    \setlength{\tabcolsep}{3pt}
    \renewcommand{\arraystretch}{1.15}
    \begin{tabular*}{\columnwidth}{@{}l@{\extracolsep{\fill}}lcrrr@{}}
        \toprule
        \textbf{Model} & \textbf{Fusion} & \textbf{Refresh}
        & \textbf{PSNR}$\uparrow$
        & \textbf{IR}$\uparrow$
        & \textbf{CLIP}$\uparrow$ \\
        \midrule
        FLUX.1-dev
        & C/T & None & 22.4570 & 0.9842 & 27.4709 \\
        & C/H & None & 22.5882 & 0.9821 & 27.4328 \\
        & C/T & C/T & 22.4893 & \textbf{0.9898} & \textbf{27.5715} \\
        & C/H & C/H & \textbf{22.6208} & 0.9822 & 27.4391 \\
        \midrule
        SD3.5-Large
        & C/T & None & 15.0729 & \textbf{0.8993} & 28.1141 \\
        & C/H & None & 15.3287 & 0.8628 & 28.2868 \\
        & C/T & C/T & 15.0924 & 0.8780 & 28.1072 \\
        & C/H & C/H & \textbf{15.7570} & 0.8987 & \textbf{28.3380} \\
        \bottomrule
    \end{tabular*}
\end{table}

As shown in Table~\ref{tab:taylor_replacement}, without adaptive refresh, C/H fusion improves PSNR over C/T by 0.1312~dB on FLUX.1-dev and 0.2558~dB on SD3.5-Large. However, C/T obtains higher IR on both models, while the CLIP preference varies across models.

With adaptive refresh, the formal C/H configuration exceeds C/T in PSNR by 0.1315~dB on FLUX.1-dev and 0.6646~dB on SD3.5-Large. On FLUX.1-dev, C/T achieves higher IR and CLIP. On SD3.5-Large, refreshed C/H outperforms refreshed C/T on all three metrics, although C/T without refresh retains a marginally higher IR. These results show a consistent PSNR advantage for Hermite fusion in the evaluated configurations, with metric-dependent trade-offs in IR and CLIP.

\subsection{Signal-Layer Ablation on SD3.5-Large}
\label{sec:app-sd3-layer}

We further investigate the effect of the disagreement signal layer on SD3.5-Large at NFE$=10$. We compare six 1-based joint-block indices, $p\in\{1,10,19,20,29,38\}$. Each layer receives an independently calibrated Q90 threshold, estimated from post-warmup Cache-step signals on the same scheduled development trajectories, with refresh disabled during calibration. All thresholds are frozen before evaluation on the disjoint DrawBench200 test set.

Table~\ref{tab:app-sd3-layer} reports mean per-image PSNR against a matched Full50 reference, ImageReward (IR), and CLIP scores over all 200 test prompts. The preferred signal layer depends on the evaluation metric. Layer 29 achieves the highest PSNR among the six candidates, exceeding layers 1 and 38 by 0.231~dB and 0.037~dB, respectively. In contrast, layer 19 yields the highest ImageReward, while layer 20 achieves the highest CLIP score. Thus, intermediate-layer disagreement can provide a useful alternative to the final-layer signal, but no tested layer maximizes all metrics.

\begin{table}[t]
\centering
\small
\caption{Signal-layer ablation on SD3.5-Large at
NFE$=10$, averaged over 200 DrawBench prompts.
Joint-block indices are 1-based. Each layer uses
an independently calibrated Q90 threshold.
Bold indicates the best result in each column.}
\label{tab:app-sd3-layer}
\renewcommand{\arraystretch}{1.12}
\begin{tabular*}{\columnwidth}
{@{}c@{\extracolsep{\fill}}ccc@{}}
\toprule
Layer & PSNR$\uparrow$ & IR$\uparrow$ & CLIP$\uparrow$ \\
\midrule
1  & 15.645 & 0.8939 & 28.277 \\
10 & 15.656 & 0.9000 & 28.291 \\
19 & 15.715 & \textbf{0.9151} & 28.336 \\
20 & 15.769 & 0.9070 & \textbf{28.348} \\
29 & \textbf{15.876} & 0.9015 & 28.303 \\
38 & 15.839 & 0.8961 & 28.315 \\
\bottomrule
\end{tabular*}
\end{table}

\subsection{Horizon-Conditioned versus Global Feedback}
\label{sec:app-feedback-scope}

We compare horizon-conditioned feedback (Horizon) with globally shared feedback (Global) on FLUX.1-dev at NFE budgets of 18, 14, 10, 9, and 8. Horizon maintains separate C/H fusion weights for different forecast horizons and attenuates feedback transfer with horizon distance. Global maintains a single shared fusion weight per sample across all forecast horizons and updates it using an exponential moving average of the same feedback targets. Both variants update weights only at eligible Full evaluations; Cache steps provide no ground-truth feedback.

Table~\ref{tab:app-feedback-scope} reports mean PSNR, ImageReward (IR), and CLIP scores over 200 DrawBench prompts. Horizon achieves slightly higher PSNR at NFE$=18$, 14, and 10, with the largest gain of 0.049~dB at NFE$=14$. At NFE$=9$ and 8, Global instead achieves higher PSNR by 0.070~dB and 0.033~dB, respectively, while Horizon yields higher IR and CLIP scores. Across all five budgets, Horizon has higher IR in four cases and higher CLIP in three cases, although the differences are small. These results indicate that horizon conditioning changes the balance between reference fidelity and text-image quality, without providing a uniform advantage across all budgets and metrics.

\begin{table}[t]
    \centering
    \small
    \caption{Horizon-conditioned versus global feedback on
    FLUX.1-dev across five NFE budgets, averaged over
    200 DrawBench prompts. Bold denotes the better
    result within each budget.}
    \label{tab:app-feedback-scope}
    \renewcommand{\arraystretch}{1.12}
    \begin{tabular*}{\columnwidth}{@{}c@{\extracolsep{\fill}}cccc@{}}
        \toprule
        NFE & Feedback & PSNR$\uparrow$ & IR$\uparrow$ & CLIP$\uparrow$ \\
        \midrule
        18 & Global  & 25.657 & 0.9866 & \textbf{27.388} \\
           & Horizon & \textbf{25.664} & \textbf{0.9900} & 27.379 \\
        \midrule
        14 & Global  & 24.837 & 0.9760 & \textbf{27.341} \\
           & Horizon & \textbf{24.886} & \textbf{0.9787} & 27.339 \\
        \midrule
        10 & Global  & 22.616 & \textbf{0.9826} & 27.421 \\
           & Horizon & \textbf{22.621} & 0.9822 & \textbf{27.439} \\
        \midrule
         9 & Global  & \textbf{21.954} & 0.9634 & 27.473 \\
           & Horizon & 21.884 & \textbf{0.9638} & \textbf{27.529} \\
        \midrule
         8 & Global  & \textbf{20.252} & 0.8797 & 27.687 \\
           & Horizon & 20.219 & \textbf{0.8822} & \textbf{27.713} \\
        \bottomrule
    \end{tabular*}
\end{table}

\section{Theoretical Analysis}
\label{sec:app-theory}

We analyze the approximation error of Chebyshev-Hermite fusion,
the horizon-conditioned feedback rule, and the effect of local
feature errors on sampling.

\subsection{Notation and Fusion Error}
\label{sec:app-theory-notation}

Following the main paper, $C_t^{(L)}$, $H_t^{(L)}$, and
$F_t^{(L)}$ denote the Chebyshev forecast, Hermite forecast,
and actual feature at the cached-output layer $L$. At Cache
steps, $F_t^{(L)}$ is a hypothetical Full feature at the current
accelerated state, unavailable to the algorithm. Define
\begin{equation}
 e_{i,t}=\frac{P_{i,t}^{(L)}-F_t^{(L)}}
 {\sqrt{\|F_t^{(L)}\|_2^2+\epsilon_e}},\quad i\in\{C,H\},
 \label{eq:app-theory-residuals}
\end{equation}
where $P_{C,t}^{(L)}=C_t^{(L)}$ and
$P_{H,t}^{(L)}=H_t^{(L)}$, so $E_{i,t}=\|e_{i,t}\|_2^2$.
For a Hermite weight $w\in[0,1]$, expansion of the squared
norm gives the exact identity
\begin{equation}
\begin{aligned}
 \mathcal E_t(w)
 &=\|(1-w)e_{C,t}+we_{H,t}\|_2^2\\
 &=(1-w)E_{C,t}+wE_{H,t}\\
 &\quad-w(1-w)\|e_{C,t}-e_{H,t}\|_2^2.
\end{aligned}
\label{eq:app-theory-fusion-identity}
\end{equation}
Thus fusion does no worse than the corresponding weighted
average of individual squared errors. This does \emph{not}
ensure that it outperforms the better individual forecaster.
For instance, if one predictor has much smaller error,
a poorly chosen mixture weight can degrade its performance.
Thus the identity motivates exploiting complementary residuals,
but accurate selection of the mixture weight remains essential.

\subsection{Horizon-Conditioned Optimality}
\label{sec:app-theory-optimality}

Let $\mathcal R_h(w)=\mathbb E[\mathcal E_t(w)\mid h_t=h]$
under a fixed forecast distribution, and define
\begin{equation}
\begin{aligned}
 A_h&=\mathbb E[\|e_{C,t}\|_2^2\mid h_t=h],\\
 B_h&=\mathbb E[\|e_{H,t}\|_2^2\mid h_t=h],\\
 G_h&=\mathbb E[\langle e_{C,t},e_{H,t}\rangle\mid h_t=h].
\end{aligned}
\label{eq:app-theory-stats}
\end{equation}

\noindent\textbf{Proposition 1 (Optimal horizon-conditioned fusion).}
For $D_h=A_h+B_h-2G_h>0$, the optimal convex weight is
\begin{equation}
 w_h^\star=\Pi_{[0,1]}\!\left(\frac{A_h-G_h}{D_h}\right),
 \label{eq:app-theory-optimum}
\end{equation}
where $\Pi_{[0,1]}$ denotes clipping to $[0,1]$.
If $D_h=0$, both forecasts coincide almost surely.
Let $\pi_h$ be the fixed horizon frequencies and
$w_{\mathrm g}^\star$ the optimal shared weight. Then
\begin{equation}
 \sum_h\pi_h\mathcal R_h(w_h^\star)
 \leq\sum_h\pi_h\mathcal R_h(w_{\mathrm g}^\star).
 \label{eq:app-theory-oracle}
\end{equation}
\noindent\emph{Proof.}
Expanding gives
$\mathcal R_h(w)=A_h+2(G_h-A_h)w+D_hw^2$;
minimizing this convex quadratic proves
Eq.~\eqref{eq:app-theory-optimum}. When $D_h=0$,
$\mathbb E[\|e_{C,t}-e_{H,t}\|_2^2\mid h_t=h]=0$.
The inequality follows because independent horizon weights
can reproduce any globally shared weight. \hfill$\square$

When all horizon-specific optima are interior, the excess
risk of the globally shared optimum is exactly
$\sum_h\pi_hD_h(w_{\mathrm g}^\star-w_h^\star)^2$.
The inequality may be strict whenever the preferred weights
differ across horizons with positive $D_h$ and $\pi_h$.
This is an \emph{oracle} comparison: CH-Cache learns weights
from sparse Full-step feedback, rather than knowing the
Cache-step errors. In particular, the optimal weight depends
on $G_h$, which the implemented feedback does not estimate.

\subsection{Feedback Target and Cross-Horizon Stability}
\label{sec:app-theory-feedback}

At a valid Full evaluation, CH-Cache uses
$v_t=(E_{C,t}+\epsilon_e)/(E_{C,t}+E_{H,t}+2\epsilon_e)$.
This target is the unique minimizer over $[0,1]$ of
\begin{equation}
 J_t(w)=(1-w)^2(E_{C,t}+\epsilon_e)
       +w^2(E_{H,t}+\epsilon_e),
 \label{eq:app-theory-surrogate}
\end{equation}
as verified by $J_t'(v_t)=0$ and
$J_t''(w)=2(E_{C,t}+E_{H,t}+2\epsilon_e)>0$.
Since $\mathcal E_t(w)\leq2J_t(w)$, substituting its
minimizer also yields
\begin{equation}
 \mathcal E_t(v_t)\leq
 \frac{2(E_{C,t}+\epsilon_e)(E_{H,t}+\epsilon_e)}
 {E_{C,t}+E_{H,t}+2\epsilon_e}.
 \label{eq:app-theory-surrogate-bound}
\end{equation}
This is a bound on the surrogate-based target, not a
claim that $v_t$ minimizes the exact fusion error,
which also depends on residual correlation. In particular,
feedback is computed using the pre-update predictions
at valid Full steps and affects only subsequent Cache steps.

The horizon-indexed bank follows the main-paper update
\begin{equation}
\begin{aligned}
 \eta_t(h)&=\frac{\beta}{1+((h-h_t)/r)^2},\\
 w_{t+1}[h]&=(1-\eta_t(h))w_t[h]+\eta_t(h)v_t,
\end{aligned}
\label{eq:app-theory-cauchy}
\end{equation}
for valid feedback events; otherwise it remains unchanged.
Here $\beta\in[0,1]$ and $r>0$.

\noindent\textbf{Proposition 2 (Stable transfer and bias).}
If $w_0[h]\in[0,1]$, then $w_t[h]\in[0,1]$ for all $t$, and
$|w_{t+1}[h]-w_t[h]|\leq\eta_t(h)$.
Suppose a reference reliability function $\mu(h)\in[0,1]$
is $L_\mu$-Lipschitz across horizons. With
$\varepsilon_t=|v_t-\mu(h_t)|$,
\begin{equation}
\begin{aligned}
 |w_{t+1}[h]-\mu(h)|
 &\leq(1-\eta_t(h))|w_t[h]-\mu(h)|\\
 &\quad+\eta_t(h)(\varepsilon_t+L_\mu|h-h_t|),
\end{aligned}
\label{eq:app-theory-transfer-bound}
\end{equation}
where the horizon-mismatch term satisfies
\begin{equation}
 \eta_t(h)L_\mu|h-h_t|\leq\frac{\beta L_\mu r}{2}.
 \label{eq:app-theory-bias}
\end{equation}
\noindent\emph{Proof.}
The update is a convex combination of two values in $[0,1]$,
which gives interval preservation and the update bound.
Subtract $\mu(h)$ from Eq.~\eqref{eq:app-theory-cauchy},
apply the triangle inequality, and use
$|v_t-\mu(h)|\leq\varepsilon_t+L_\mu|h-h_t|$.
Finally, $r^2+(h-h_t)^2\geq2r|h-h_t|$ establishes
Eq.~\eqref{eq:app-theory-bias}. \hfill$\square$

Thus cross-horizon transfer shares sparse observations but
may introduce bias when reliability varies across horizons.
In the limiting cases, $r\to0^+$ concentrates updates
on the observed integer horizon, while $r\to\infty$
produces approximately uniform update rates. The latter
coincides with a global exponential average if all horizon
weights share the same initialization and feedback events.
The reference function $\mu$ is assumed fixed over the
updates being analyzed; nonstationarity would introduce
an additional drift term. None of these bounds guarantees
an improvement in image quality.

\subsection{Disagreement and Budget Preservation}
\label{sec:app-theory-control}

At signal layer $p$, let $P_{C,t}^{(p)}=C_t^{(p)}$
and $P_{H,t}^{(p)}=H_t^{(p)}$. The triangle inequality yields
\begin{equation}
\begin{aligned}
 &\max_{i\in\{C,H\}}\|P_{i,t}^{(p)}-F_t^{(p)}\|_2\\
 &\qquad\geq\tfrac12\|C_t^{(p)}-H_t^{(p)}\|_2.
\end{aligned}
\label{eq:app-theory-disagreement}
\end{equation}
Therefore, large disagreement implies that at least one
forecaster is inaccurate at layer $p$. The converse fails:
both forecasts may agree while sharing a large error.
Consequently, the normalized signal $r_t^{(p)}$ is a
heuristic refresh indicator, not a bound on fused error
at the cached-output layer $L$. No cross-layer error
guarantee follows without assumptions relating $p$ and $L$.

For Refresh/Repay, let $N=|\mathcal S_0|$ and let $R$
and $P$ count accepted refreshes and repayments. The
realized Full count is $N_{\mathrm{actual}}=N+R-P$.
If each refresh is matched by an eligible future repayment,
then $R=P$ and $N_{\mathrm{actual}}=N$.
This equality assumes that protected Full steps, including
warmup, are never counted as available repayment slots.
The result guarantees an evaluation count, not equal runtime,
trajectory accuracy, or image quality across schedules.

\subsection{Propagation of Feature Approximation Error}
\label{sec:app-theory-trajectory}

Let $x_t$ and $x_t^\star$ be accelerated and all-Full
states with identical initialization. Let $\Phi_t$ and
$\widehat\Phi_t$ denote Full and accelerated sampling updates.
Assume
\begin{equation}
\begin{aligned}
 \|\Phi_t(x)-\Phi_t(y)\|_2&\leq\Lambda_t\|x-y\|_2,\\
 \|\widehat\Phi_t(x_t)-\Phi_t(x_t)\|_2
 &\leq K_t\|\Delta_t\|_2,
\end{aligned}
\label{eq:app-theory-sampler-assumptions}
\end{equation}
where $\Delta_t=\widehat F_t-F_t^{(L)}$ is the
unnormalized feature defect at a Cache step, and
$\Delta_t=0$ at an actual Full step.
Writing $\zeta_t=\|x_t-x_t^\star\|_2$, the triangle inequality
gives $\zeta_{t+1}\leq\Lambda_t\zeta_t+K_t\|\Delta_t\|_2$.
Unrolling from $\zeta_0=0$ yields
\begin{equation}
 \|x_T-x_T^\star\|_2
 \leq\sum_{t\in\mathcal C}K_t\|\Delta_t\|_2
 \prod_{s=t+1}^{T-1}\Lambda_s,
 \label{eq:app-theory-trajectory}
\end{equation}
where $\mathcal C$ is the realized Cache-step set.
If the decoder is $L_D$-Lipschitz, the decoded-image
difference is at most $L_D$ times the right-hand side
of Eq.~\eqref{eq:app-theory-trajectory}.
The result is conditional on the sampler and local-defect
bounds: a subsequent Full step removes its own local
approximation defect but does not reset the deviations
accumulated at earlier Cache steps. It does not establish
a bound on perceptual evaluation metrics.

\begin{figure*}[!t]
    \centering
    \includegraphics[width=\textwidth]
    {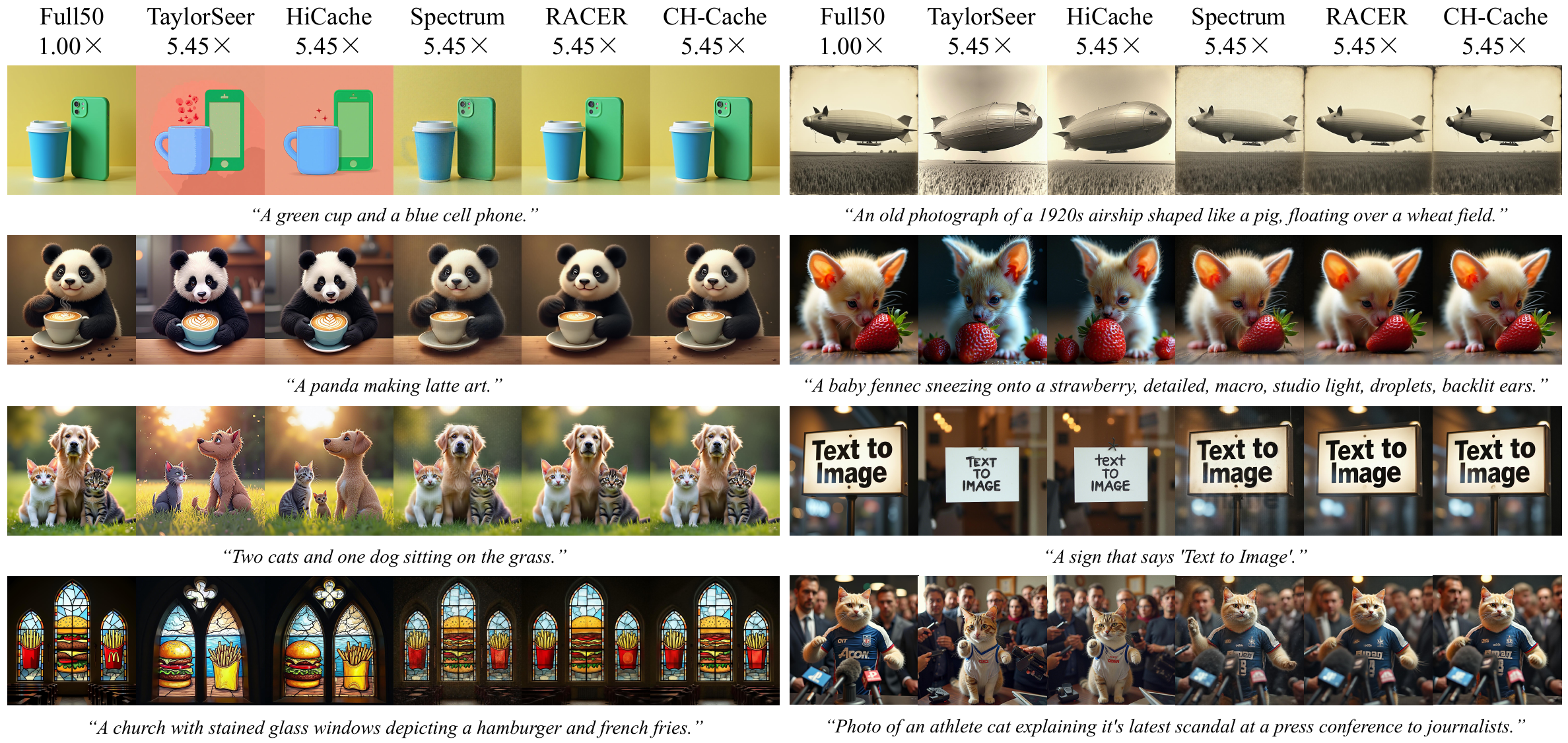}
    \caption{Qualitative comparisons on FLUX.1-dev
    at NFE$=9$ for all accelerated methods.}
    \label{fig:app-qual-flux}
\end{figure*}

\section{Qualitative Results}
\label{sec:app-qualitative}

We present additional qualitative comparisons on
FLUX.1-dev and SD3.5-Large using representative
DrawBench prompts. For each model, all accelerated
methods are evaluated under the same number of
function evaluations and compared with the
corresponding Full50 reference. We focus on reference
fidelity in object attributes, spatial relationships,
local structure, and text rendering rather than
visual appeal alone.

\subsection{Qualitative Comparisons on FLUX.1-dev}
\label{sec:app-qual-flux}

Figure~\ref{fig:app-qual-flux} illustrates object
arrangements, character appearance, complex scenes,
and text rendering. In the cup-and-phone example,
CH-Cache retains the reference colors and positions,
whereas TaylorSeer and HiCache introduce a handled mug
and rearrange the objects. For the latte-art scene,
CH-Cache follows the panda's pose and cup composition.
It also preserves the three animals and their arrangement
in the cat-and-dog scene, as well as the window layout
and food motifs in the stained-glass example. Spectrum
and RACER remain close to Full50 in several cases.

The remaining cases emphasize unusual objects, text,
and multi-person scenes. CH-Cache preserves the
pig-shaped airship's viewpoint and silhouette and
retains the fennec, strawberry, and warm lighting,
while TaylorSeer and HiCache show larger changes in
shape or composition. It also reproduces the
``Text to Image'' sign's lettering and layout.
In the athlete-cat press-conference example, CH-Cache
maintains the cat's sports jersey, foreground position,
and surrounding journalists, whereas TaylorSeer and
HiCache alter the subject or scene arrangement.
Spectrum and RACER are also close to the reference
in this example. These comparisons illustrate
reference-specific fidelity in the selected images.

\begin{figure}[!t]
    \centering
    \includegraphics[width=\columnwidth]
    {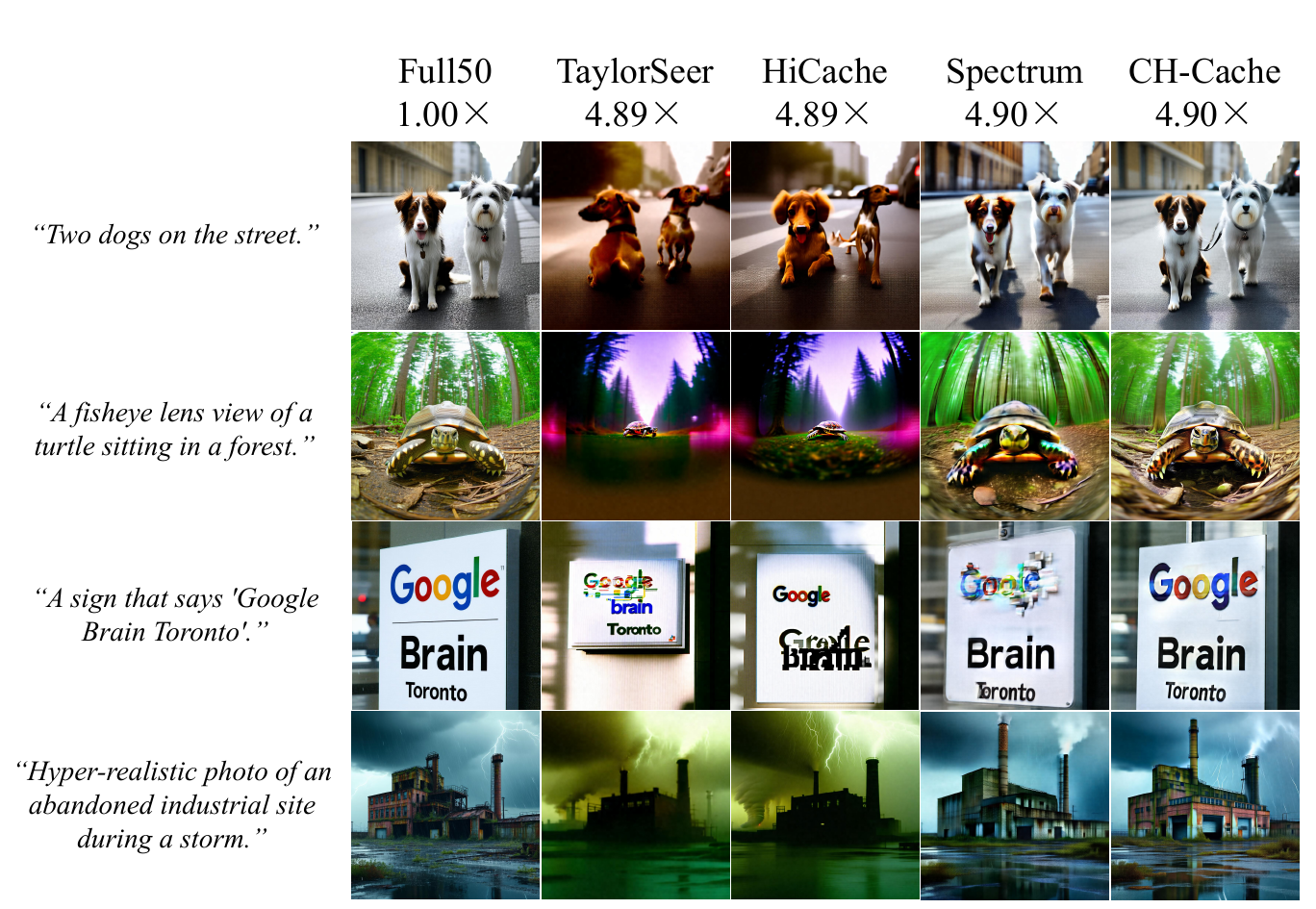}
    \caption{Qualitative comparisons on SD3.5-Large at NFE$=10$ for all accelerated
    methods.}
    \label{fig:app-qual-sd35}
\end{figure}

\subsection{Qualitative Comparisons on SD3.5-Large}
\label{sec:app-qual-sd35}

Figure~\ref{fig:app-qual-sd35} provides complementary
examples on SD3.5-Large. In the two-dog scene,
CH-Cache preserves the subjects' contrasting
appearances and relative positions, whereas
TaylorSeer and HiCache alter their appearance
and the surrounding scene. In the fisheye forest
view, CH-Cache retains the large foreground turtle
and the woodland perspective; TaylorSeer and
HiCache instead depict a much smaller subject
against a markedly different background.
The sign example further highlights text rendering:
CH-Cache reproduces the multi-line
``Google Brain Toronto'' layout more faithfully,
while the other predictions exhibit varying degrees
of text and layout distortion. 

In the industrial scene, CH-Cache maintains
recognizable buildings, chimneys, wet ground,
and stormy lighting. However, some architectural
details and facade colors still differ from
the Full50 reference, and Spectrum also preserves
much of the original composition. Together, these
examples show the benefits of CH-Cache for selected
spatial and textual details.

\begin{figure}[!t]
    \centering
    \includegraphics[width=\columnwidth]
    {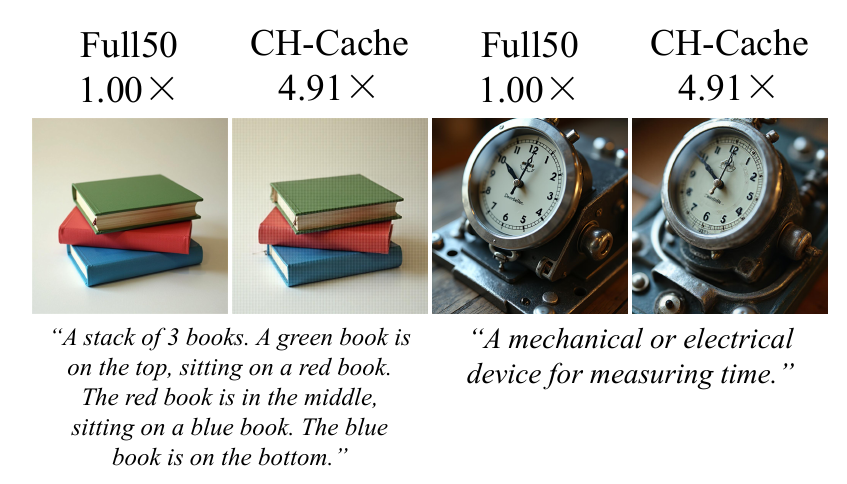}
    \caption{Limitations on FLUX.1-dev.}
    \label{fig:app-limit-flux}
\end{figure}

\section{Limitations}
\label{sec:app-limitations}

Although CH-Cache preserves the main composition of
the Full50 reference, fine details may change under
feature reuse. As illustrated in
Fig.~\ref{fig:app-limit-flux}, CH-Cache maintains the
green--red--blue ordering and approximate arrangement
of the three stacked books. However, visible grid-like
patterns appear on their surfaces, and the book edges
and nearby background differ from the reference. For the mechanical
timekeeping device, CH-Cache retains the circular
dial and metallic housing, while some numerals,
tick marks, and small components differ from Full50.
These cases show that scene-level fidelity does not
necessarily ensure accurate reconstruction of local
textures and fine geometry.

\end{document}